\documentclass[letterpaper]{article} 
\usepackage{aaai2026}  
\usepackage{times}  
\usepackage{helvet}  
\usepackage{courier}  
\usepackage[hyphens]{url}  
\usepackage{graphicx} 
\usepackage{natbib}  
\usepackage{caption} 
\usepackage{algorithm}
\usepackage{algorithmic}
\usepackage{array}
\usepackage{ragged2e}

\newfloat{code}{H}{lop}
\usepackage[hyphens]{url}
\usepackage{graphicx}
\usepackage[table]{xcolor}
\usepackage{colortbl} 
\usepackage{natbib}
\usepackage{caption}
\usepackage{multirow}
\usepackage{booktabs}
\usepackage{amsmath}

\usepackage{newfloat}
\usepackage{listings}
\DeclareCaptionStyle{ruled}{labelfont=normalfont,labelsep=colon,strut=off} 
\floatstyle{ruled}
\newfloat{listing}{tb}{lst}{}
\floatname{listing}{Listing}
\title{CrossCheck-Bench: Diagnosing Compositional Failures \\ in Multimodal Conflict Resolution}
\author{
    Baoliang Tian\textsuperscript{\rm 1}\equalcontrib, 
    Yuxuan Si\textsuperscript{\rm 1,\rm 2}\equalcontrib, 
    Jilong Wang\textsuperscript{\rm 1,\rm 3}\equalcontrib, 
    Lingyao Li\textsuperscript{\rm 1}, 
    Zhongyuan Bao\textsuperscript{\rm 1}, 
    Zineng Zhou\textsuperscript{\rm 1}, \\
    Tao Wang\textsuperscript{\rm 1{\rm \textsuperscript{\dag}}}, 
    Sixu Li\textsuperscript{\rm 1}, 
    Ziyao Xu\textsuperscript{\rm 1}, 
    Mingze Wang\textsuperscript{\rm 1}, 
    Zhouzhuo Zhang\textsuperscript{\rm 1}, 
    Zhihao Wang\textsuperscript{\rm 1}, \\
    Yike Yun\textsuperscript{\rm 1}, 
    Ke Tian\textsuperscript{\rm 1}, 
    Ning Yang\textsuperscript{\rm 3}\thanks{Corresponding author.}, 
    Minghui Qiu\textsuperscript{\rm 1}
}
\affiliations{
    \textsuperscript{\rm 1}ByteDance\\
    \textsuperscript{\rm 2}Zhejiang University\\
    \textsuperscript{\rm 3}Institute of Automation, Chinese Academy of Sciences\\
    walton.wang929@gmail.com {\rm {\rm \textsuperscript{\dag}}}, ning.yang@ia.ac.cn {\rm {\rm \textsuperscript{\dag}}}
}

\usepackage{bibentry}

\begin{document}

\maketitle

\begin{abstract}
Multimodal Large Language Models are primarily trained and evaluated on aligned image-text pairs, which leaves their ability to detect and resolve real-world inconsistencies largely unexplored. In open-domain applications visual and textual cues often conflict, requiring models to perform structured reasoning beyond surface-level alignment.
We introduce CrossCheck-Bench, a diagnostic benchmark for evaluating contradiction detection in multimodal inputs. The benchmark adopts a hierarchical task framework covering three levels of reasoning complexity and defines seven atomic capabilities essential for resolving cross-modal inconsistencies. CrossCheck-Bench includes 15k question-answer pairs sourced from real-world artifacts with synthetically injected contradictions. The dataset is constructed through a multi-stage annotation pipeline involving more than 450 expert hours to ensure semantic validity and calibrated difficulty across perception, integration, and reasoning.
We evaluate 13 state-of-the-art vision-language models and observe a consistent performance drop as tasks shift from perceptual matching to logical contradiction detection. Most models perform well on isolated entity recognition but fail when multiple clues must be synthesized for conflict reasoning. Capability-level analysis further reveals uneven skill acquisition, especially in tasks requiring multi-step inference or rule-based validation.
Additional probing shows that conventional prompting strategies such as Chain-of-Thought and Set-of-Mark yield only marginal gains. By contrast, methods that interleave symbolic reasoning with grounded visual processing achieve more stable improvements. These results highlight a persistent bottleneck in multimodal reasoning and suggest new directions for building models capable of robust cross-modal verification.
\end{abstract}

\begin{links}
    \link{Code}{https://github.com/bytedance/CrossCheck-Bench}
\end{links}

\begin{figure}[ht]
  \centering
  \includegraphics[width=\linewidth]{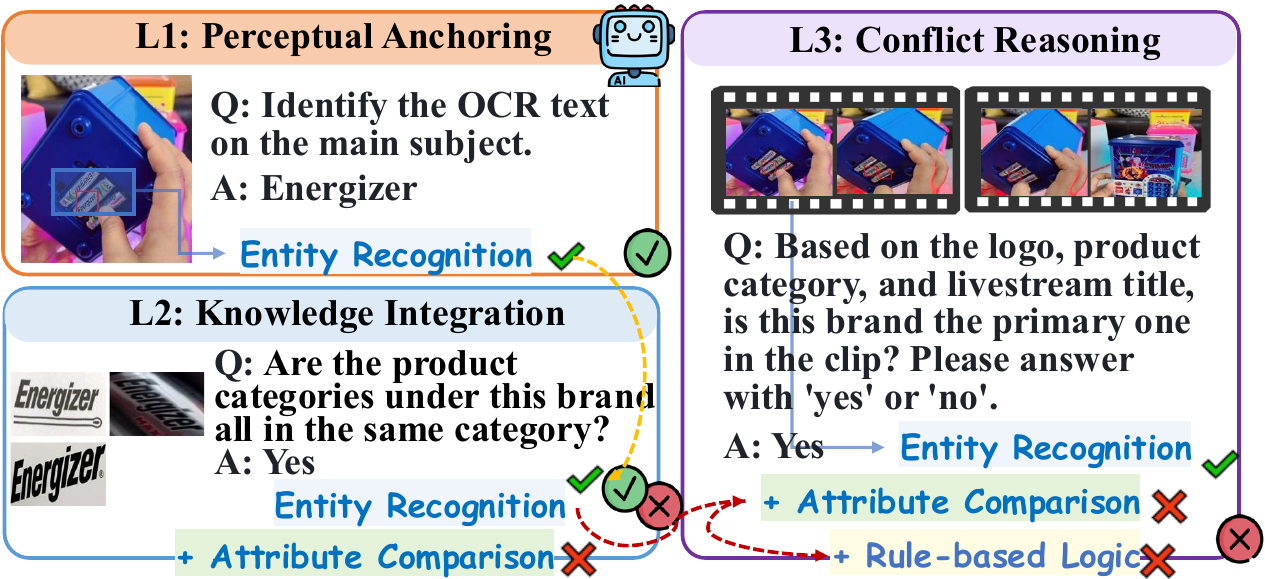}
  \caption{\textsc{CrossCheck-Bench} cascade case: the model answers the Level-1 perception query correctly, yet fails the dependent Level-2 integration and Level-3 conflict-reasoning tasks.}
  \label{fig:task-overview}
\end{figure}

\begin{figure*}[ht]  
    \centering
    \includegraphics[width=1\textwidth]{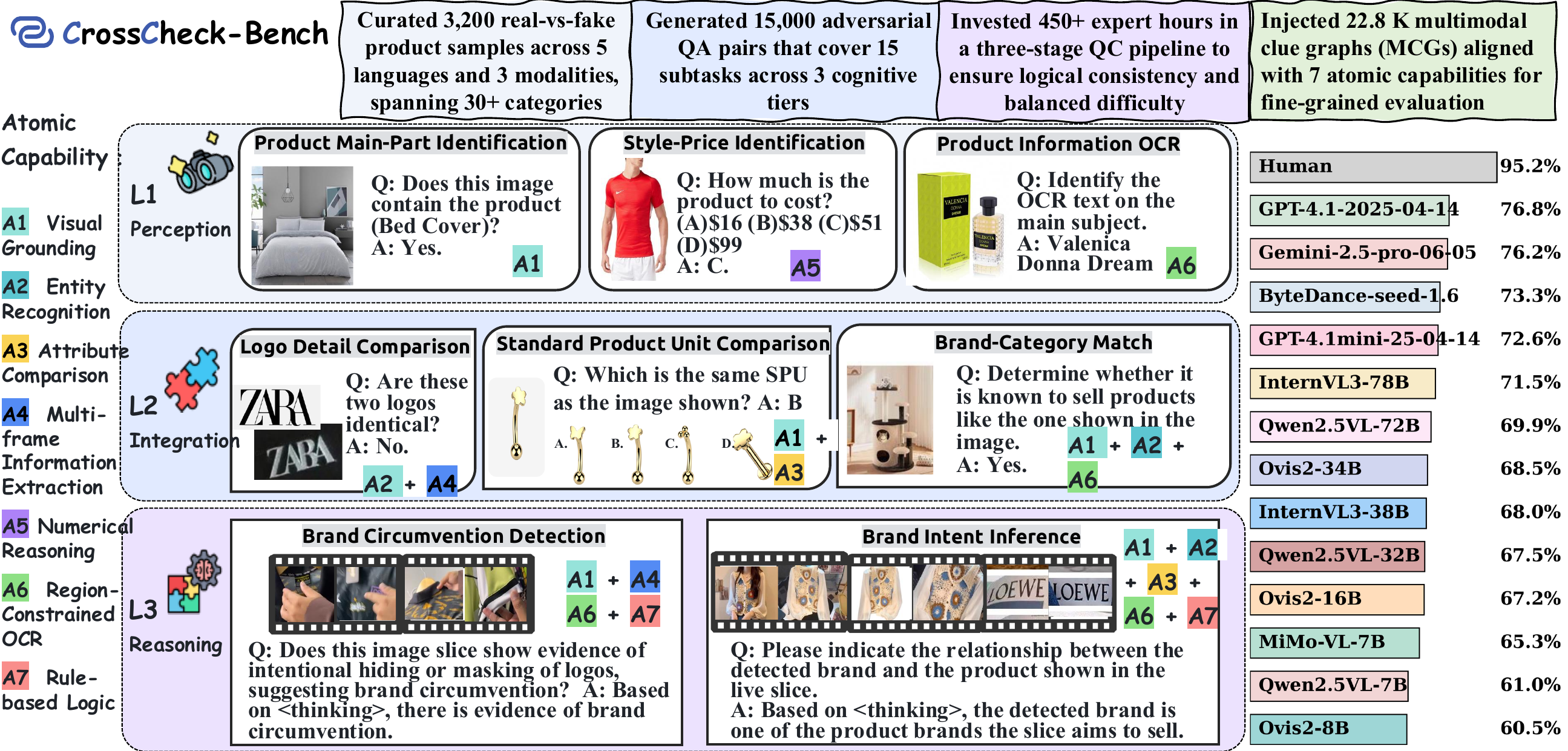}
    \caption{Overview of \textsc{CrossCheck-Bench}.  
The benchmark spans three cognitive tiers—L1 Perception, L2 Integration, and L3 Reasoning—grounded in seven atomic capabilities (A1–A7) and eight representative tasks.  
It offers thousands product samples, 14{,}690 adversarial QA pairs, and 22.8 K multimodal clue graphs, curated through 450+ expert hours.
The right-hand bar chart contrasts average model accuracy with the human upper bound.}
    \label{fig:benchmark_overview}
\end{figure*}

\section{Introduction}
Multimodal content in the open world frequently exhibits noisy, unreliable, and even deceptive characteristics. A product page may display a luxury brand logo with a suspiciously low price, or pair an image of a sports shoe with a textual description of formal wear.
Humans can instinctively recognize when such visual and textual clues do not align, flagging potential fraud or misrepresentation. 
This ability to resolve cross-modal conflicts is fundamental for any robust multimodal reasoning system~\cite{wang2023aligning,wolf2023fundamental}.
However, vision-language models (VLMs), which now underpin many content understanding deployments, have not been thoroughly evaluated on their ability to detect and reject contradictions~\cite{tan2020detecting}.
Most existing VLMs~\cite{li2023blip, alayrac2022flamingo, liu2023visual} are primarily trained and evaluated on aligned datasets, where vision and language describe the same semantic content.
This alignment-centric paradigm encourages cross-modal consistency but overlooks a critical question: \textit{can models verify whether multimodal signals are logically compatible?} 
The lack of this capability would pose a tangible risk: models may confidently affirm incompatible clues, producing outputs that are not only inaccurate but also logically inconsistent with the input evidence~\cite{li2023evaluating, liu2023trustworthy}.

Consequently, there is an urgent need for a benchmark that can effectively evaluate such an ability~\cite{ma2023crepe}.
While existing benchmarks effectively assess tasks like retrieval~\cite{yang2022blip, wasserman2025real}, description~\cite{maaz2023video, yue2024mmmu}, and entailment~\cite{xie2019visual}, few explicitly test whether multimodal inputs are jointly compatible in a structured, diagnostic way.
Unlike generation correctness evaluation, validating conflict resolution requires more than local grounding or factual recall—it demands compositional verification across modalities~\cite{johnson2017clevr}.
Specifically, this ability requires models to (1) localize corresponding semantic entities or attributes in each modality and (2) assess whether these aligned clues are logically compatible.
However, constructing such data is highly challenging, as it depends on reliable data sources to identify contradictory entities across different modalities and create realistic conflict-based evaluation tasks~\cite{johnson2017clevr}.

To address this gap, we introduce \textit{CrossCheck-Bench}, the benchmark designed to evaluate and diagnose VLMs' capacity to resolve multimodal inconsistencies~\cite{li2024survey,bai2024hallucination,gunjal2024detecting}.
As illustrated in Figure~\ref{fig:benchmark_overview}, we structure the benchmark into a three-level hierarchy reflecting increasing reasoning complexity~\cite{yue2024mmmu,lu2023mathvista}. {L1} ({Perceptual Anchoring}) evaluates whether the model can extract atomic entities from each modality. {L2} ({Knowledge Integration}) assesses whether the model can compare cross-modal attributes. {L3} ({Conflict Reasoning}) requires the model to detect implicit contradictions that arise from multi-attribute combinations. Each level builds on prior capabilities: from recognizing entities, to comparing attributes, to applying rule-based logic under uncertainty.
To enable fine-grained diagnosis, we further decompose the evaluation into seven atomic capabilities spanning Entity Recognition, Attribute Comparison, Numerical Reasoning, Multi-Frame Extraction, {etc}. 
By structuring both tasks and skills hierarchically, CrossCheck-Bench reveals how early-stage failures—such as misidentifying logos or misreading text—can propagate upward, leading to confident but flawed high-level inferences. 
As exemplified by the case study in Figure~\ref{fig:task-overview}, a single input sample may trigger success at L1 while failing at L2 and L3, illustrating how surface-level perception often masks deeper reasoning collapse.
Each QA sample is derived from real-world stimuli (e.g., e-commerce listings, ads, social posts) with injected inconsistencies requiring nontrivial inference.

We evaluated 13 current top performing VLMs, including GPT-4.1~\cite{achiam2023gpt}, Gemini-2.5~\cite{comanici2025gemini}, Qwen2.5-VL~\cite{bai2025qwen2}, InternVL3~\cite{zhu2025internvl3}, and MiMo-VL~\cite{xiaomi2025mimo}, on 15k conflict-oriented QA samples spanning all three task levels. 
Our results reveal a stark performance drop from L1 to L3: while most models succeed at local grounding, nearly all fail at multi-attribute contradiction detection, affirming implausible combinations with high confidence. 
We further examine whether prompting strategies and lightweight adaptation can enhance model performance on conflict-sensitive tasks. Our findings show that conventional methods such as Chain-of-Thought prompting and visual grounding through annotated regions provide limited benefit and may even reinforce superficial patterns. In contrast, approaches that support iterative reasoning across visual and textual inputs, including lightweight supervised fine-tuning, result in more consistent improvements. Overall, these results highlight a critical bottleneck in the ability of MLLMs to perform robust, integrated reasoning—a key challenge for future research.

\begin{figure*}[ht]  
    \centering
    \includegraphics[width=\linewidth]{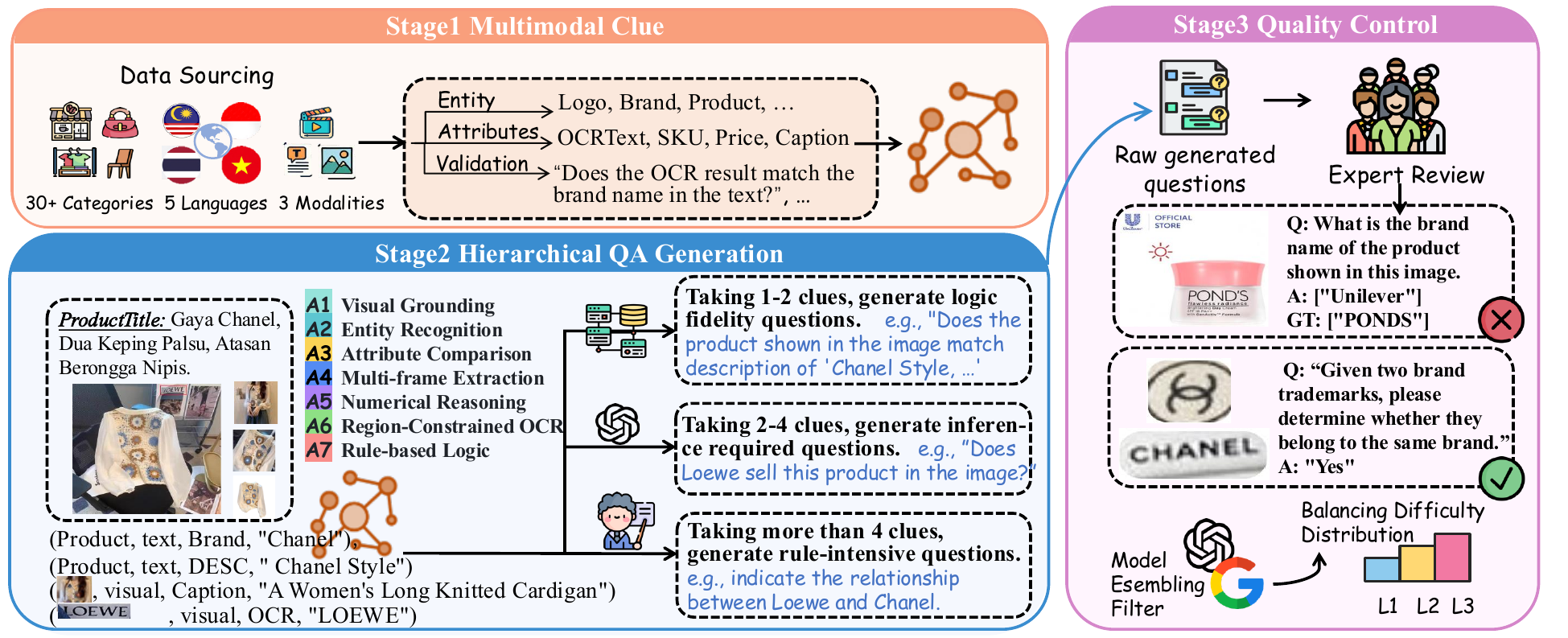}
    \caption{Dataset–construction pipeline for \textsc{CrossCheck-Bench}.  
Stage 1: Clue Encoding. Aggregates multimodal data (30+ categories, 5 languages) into clue graphs binding entities with validated attributes.
Stage 2: QA Composition. Samples $1-n$ clues to generate hierarchical QA pairs targeting 7 capabilities across 3 cognitive tiers (L1--L3). 
Stage 3: Quality Control. Employs a three-step loop (expert review, model filtering, and difficulty balancing) to ensure correctness and task uniformity.}
    \label{fig:pipeline}
\end{figure*}

\noindent To summarize, we make the following contributions:
\begin{itemize}
    \item We propose \textit{CrossCheck-Bench}, a new benchmark to evaluate and diagnose VLMs' ability to resolve multimodal inconsistencies, structured around a three-level reasoning hierarchy and seven atomic capabilities.
    \item We construct a large-scale dataset of 15k QA samples derived from real-world artifacts, with programmatically injected contradictions that require compositional reasoning to resolve.
    \item We benchmark 13 leading VLMs, revealing their systematic failure on L3 conflict reasoning, identifying the limitations of current prompting methods, and highlighting iterative verification as a promising path forward.
\end{itemize}

\section{Related Work}

\paragraph{Multimodal Reasoning Benchmarks.}
Existing benchmarks primarily evaluate VLMs on compositional tasks where modalities \textit{reinforce} each other. Datasets like VCR~\cite{zellers2019recognition}, NLVR2~\cite{suhr2019corpus}, and SNLI-VE~\cite{xie2019visual} assess textual entailment or reasoning with aligned visual support, assuming visual-textual consistency. Recent benchmarks such as MMMU~\cite{yue2024mmmu} and MathVista~\cite{lu2023mathvista} emphasize complex multimodal reasoning but remain confined to scenarios with concordant inputs. While these efforts demonstrate VLMs' growing proficiency in integrated understanding, they fail to assess models' resilience when modalities \textit{conflict}—the core challenge addressed by our work.

\paragraph{Inconsistency Detection in Vision-Language.}
Prior research on multimodal inconsistency has primarily focused on specialized, narrow tasks~\cite{elazar2021measuring,tahmasebi2024multimodal}. The MMIR benchmark~\cite{yan2025multimodal} evaluates inconsistency reasoning in layout-rich artifacts but restricts its scope to predefined error types and lacks granular diagnosis of underlying capability failures. Works like {Beyond Appearance}~\cite{xu2025toward} study modality gaps in specific attributes (e.g., color, shape) but do not scale to compositional real-world conflicts. Crucially, VLM2-Bench~\cite{zhang2025vlm2} addresses a distinct challenge: visually \textit{linking matching cues} across different images (e.g., identifying the same person), which represents an orthogonal problem to resolving \textit{contradictory multimodal evidence} within unified inputs. CrossCheck-Bench thus fills a critical void by introducing the first hierarchical evaluation framework ($\text{L1} \rightarrow \text{L3}$) to diagnose \textit{why} models fail at cross-modal conflict resolution—requiring both structural alignment and logical comparison within unified multimodal inputs.

\paragraph{Diagnostic Evaluation of VLMs.}
Efforts to diagnose VLM failures typically dissect atomic capabilities in isolation. SpaCE-10~\cite{gong2025space} decomposes spatial intelligence into 10 atomic skills but does not examine their interplay under \textit{conflicting evidence}. BEiT-3~\cite{wang2023image} and LLaVA~\cite{liu2023visual} analyze modality biases via probing tasks, yet these remain decoupled from real-world inconsistency scenarios. Our benchmark uniquely integrates capability-centric diagnosis with adversarial multimodal conflict: we not only define 7 atomic cross-checking capabilities (e.g., entity grounding, attribute verification) but deliberately stress-test their composition in failure-prone contexts where cues contradict, thereby exposing cascading reasoning breakdowns unseen in prior benchmarks.

\begin{figure}[ht!]
    \centering
    \includegraphics[width=\linewidth]{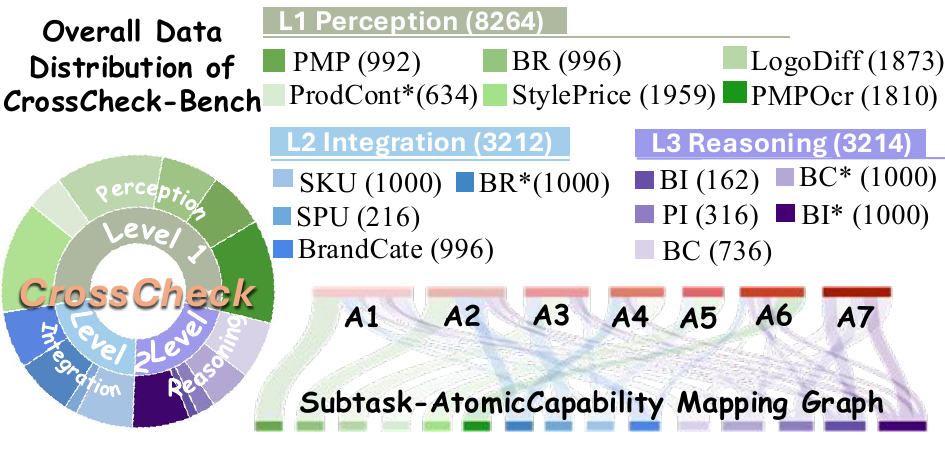}
    \caption{Dataset statistics for \textsc{CrossCheck-Bench}.  
The benchmark contains \(\sim\!15{,}000\) question–answer pairs distributed over 15 subtasks and three cognitive levels (left).  
Six subtasks probe a single atomic capability (A1–A6), while the remaining nine require compositions capabilities (right).  
Subtask names followed by “*” take multi-frame input.}
    \label{fig:data_stat}
\end{figure}

\section{CrossCheck-Bench}
\label{sec:construction}
To enable fine-grained diagnosis of multimodal inconsistency resolution in Vision-Language Models (VLMs), we introduce \textit{CrossCheck-Bench}, a factually grounded benchmark comprising 7 atomic capabilities and 15 systematically constructed tasks.
As shown in Figure~\ref{fig:benchmark_overview}, CrossCheck-Bench follows a hierarchical structure, categorizing tasks into {Perception (L1)}, {Integration (L2)}, and {Reasoning (L3)} levels, based on the combination of atomic capabilities and task complexity.
To ensure the reliability and consistency,
the construction of CrossCheck-Bench involves three key stages (see Figure~\ref{fig:pipeline}): Data Collection via Multimodal Clue Graphs, Hierarchical Tasks Generation, and Quality Verification. 
In total, \textbf{over 450 expert-hours} were invested to curate 14.69k high-fidelity, semantically grounded QA tasks.

\subsection{Data Collection}
\label{ssec:ground_truth}
We curated a diverse e-commerce dataset from major platforms, comprising 22.8k listings with $\ge 5$ verified attributes and high-resolution images. To represent factual signals, we define \textit{Multimodal Cue Graphs} (MCGs) as quadruples: \texttt{(entity, modality, attribute, value)}~\cite{kommineni2024human, yao2025exploring}. MCG construction proceeds in three stages:

Entity Extraction: We employ an ensemble of YOLOv8-L~\cite{yi2023small}, GroundingDINO~\cite{liu2024grounding}, and visual embeddings for image-level detection, alongside fine-tuned Qwen3-8B~\cite{yang2025qwen3} for textual entity recognition.
Attribute Extraction: Visual (e.g., OCR, shape) and textual (e.g., brand, price) signals are extracted via rule-based templates and GPT-4o augmentation.
Cross-Validation: GPT-4o identifies cross-modal inconsistencies. Discrepant pairs are corrected, with a 15\% manual audit yielding 98.2\% accuracy. 

The final set includes 22.8k MCGs, each containing on average 12.7 verifiable clues, with a semantic consistency rate of 97.3\% based on expert auditing.



\subsection{Hierarchical QA Generation}
\label{ssec:generation}
To enable diagnostic evaluation across varying levels of reasoning complexity, we formulate a three-tiered task taxonomy grounded in atomic visual-language capabilities. Each task is derived from structured information encoded in the MCGs, allowing precise targeting of distinct skills and their combinations.

\noindent
\textbf{The Three-Tier Diagnostic Taxonomy.}
We define seven atomic capabilities for assessing multimodal reasoning: A1 (Visual Grounding), A2 (Entity Recognition), A3 (Attribute Comparison), A4 (Multi-frame Reasoning), A5 (Numerical Plausibility), A6 (Region-Constrained OCR), and A7 (Rule-based Logic). These skills represent fundamental cross-modal abilities required to detect real-world inconsistencies.
Building on them, we construct a three-level hierarchy to measure increasing reasoning depth. {Perception (L1)} isolates a single atomic skill to evaluate basic cross-modal alignment. {Integration (L2)} combines two to three capabilities to test coordination across modalities. {Reasoning (L3)} requires synthesizing multiple clues, applying commonsense or domain knowledge, and resolving implicit contradictions. This taxonomy is both diagnostic and necessary, enabling fine-grained attribution of failures to perceptual, integrative, or reasoning limitations.

\begin{table*}[ht!]
\centering
\small
\setlength{\tabcolsep}{3.8pt}
\renewcommand{\arraystretch}{0.85}
\begin{tabular}{m{3.5cm}|c|cccccc|cccc|ccccc} 
\toprule
\multirow{2}{*}{\textbf{Models}} &
\multirow{2}{*}{\textbf{Avg.}} &
\multicolumn{6}{c|}{\textbf{Perception}} & 
\multicolumn{4}{c|}{\textbf{Integration}} & 
\multicolumn{5}{c}{\textbf{Reasoning}} \\
\cmidrule(l){3-8} \cmidrule(l){9-12} \cmidrule(l){13-17}
& & \textbf{LD} & \textbf{BR} & \textbf{PMP} & \textbf{PMO} & \textbf{PD*} & \textbf{SP}  &  \textbf{BR*} & \textbf{SPU} & \textbf{SKU} & \textbf{BDC} & \textbf{BC} & \textbf{BC*} & \textbf{BI} & \textbf{PI} & \textbf{BI*}   \\
\midrule

\multicolumn{17}{l}{\textbf{Open-Source MLLMs (Small Group)}} \\
\midrule
Qwen2.5VL-7B & 61.0 & 59.1  & 61.8 & 78.8 & 56.6 & 53.2 & 42.9 & 62.5 & 78.4 & 59.8 & 64.9 & 58.9 & 50.0 & 80.9 & 51.6 & 55.3 \\
Ovis2-8B & 60.5  & \textbf{\underline{77.9}} & 59.8  & 72.1 & 51.1 & 72.6 & 36.6 &  67.6 & 67.9 & 56.6 & 75.5 & 58.7 & 50.1 &  66.1 & 50.3 & 44.6\\
MiMo-VL-7B  & 65.3  & 66.8  & \textbf{\underline{93.0}} & 73.7 & 59.5 & 62.9 & 52.1 & 66.1 & 76.2 & 56.4 & 91.5 & 60.7 & 52.8 & 59.3  & 61.4 & 46.7 \\
\midrule

\multicolumn{17}{l}{\textbf{Open-Source MLLMs (Medium Group)}} \\
\midrule
Ovis2-16B  &  67.2 & 77.2 & 75.4 & 74.1 & 61.4 & 73.5 & 52.0 & 71.3 & 78.0 & 69.8 & 89.9 & 58.3 & 52.6 & 75.3 & 57.6 &  42.0 \\
Qwen2.5VL-32B  & 67.5  & 67.3 & 69.5 & 79.7 & 67.7 & 71.9 & 53.9 & 69.4 & 81.2 & 52.9 & 93.1 & 67.8 & 49.9 & 65.9 & 50.0 & \textbf{\underline{72.6}} \\
Ovis2-34B  & 68.5  & 67.8  & 71.6 & 72.6 & 70.2 & 72.4 & 51.0 & 69.0 & 85.3 & 66.2 & 89.8 & 58.6 & 50.7 & 63.0 & 82.7 & 55.9  \\
InternVL3-38B & 68.0  & 64.2  & 59.7 & \textbf{\underline{92.4}} & 64.2 & 71.5 & 52.2 & 71.6 & 68.4 & 67.3 & 93.4 & 58.9 & 53.9 & 63.4 & 75.6 & 63.4 \\
\midrule
\multicolumn{17}{l}{\textbf{Open-Source MLLMs (Large Group)}} \\
\midrule
Qwen2.5VL-72B  & 69.9 & 67.3  & 69.5  & 79.7 & 75.1 & 72.1 & 51.9 & 69.1 & 86.7 & 68.3 & 90.8 & 58.6 & 50.6 & 63.6 & 72.6 & 69.2 \\
InternVL3-78B  & 71.5  & 73.3  & 69.1 & 89.0 & 74.4 & 61.5 & 50.1 & 74.0 & 81.2 & 71.3 & 91.9 & 59.6 & 53.1 & 78.8 & 81.5 & 64.0 \\
\midrule

\multicolumn{17}{l}{\textbf{Closed-Source MLLMs}} \\
\midrule
GPT-4.1-mini-2025-04-14 & 72.6 & 65.8 & 85.7 & 78.1 & 79.1 & \textbf{\underline{74.8}} & 60.4 & 74.5 & 86.7 & \textbf{\underline{73.9}} & 91.8 & 67.4 & 54.0 & 74.7 & 68.4 & 54.0 \\
ByteDance-seed-1.6 & 73.3 & 62.9 & 90.3 & 76.2 & 66.2 & 59.2 & 63.2 & 71.0 & 89.5 & 73.7 & 96.2 & 63.1 & 57.8 & \textbf{\underline{85.5}} & \textbf{\underline{81.0}} & 63.8\\
GPT-4.1-2025-04-14 & \textbf{\underline{76.8}} & 68.2 & 89.9 & 78.1 & \textbf{\underline{85.3}} & 71.1 & 70.2 & 80.1 & \textbf{\underline{90.4}} & 72.1 & \textbf{\underline{98.1}} & 67.8 & 65.4 & 74.7 & 75.7 & 65.4\\
Gemini-2.5-pro-p-06-05 & 76.2 & 69.6 & 92.0 & 76.8 & 80.9 & 70.7 & \textbf{\underline{72.2}} & \textbf{\underline{83.7}} & 85.8 & 48.2 & 97.7 & \textbf{\underline{71.0}} & \textbf{\underline{66.3}} & 77.2 & \textbf{\underline{81.0}} & 70.2 \\
\midrule
\textbf{Human} & \textbf{95.2} & \textbf{94.5} & \textbf{98.1} & \textbf{96.7} & \textbf{93.2} & \textbf{88.5} & \textbf{85.6} & \textbf{92.4} & \textbf{97.8} & \textbf{89.1} & \textbf{99.5} & \textbf{85.2} & \textbf{82.1} & \textbf{94.3} & \textbf{92.8} & \textbf{88.0} \\
\bottomrule
\end{tabular}
\caption{Main results of the \textsc{CrossCheck-Bench}. The best performance is highlighted by bold and underline. Human performance in bold as an upper-bound reference. * denotes multi-frame input.}
\label{tab:overall}

\end{table*}

\noindent
\textbf{Question Generation.}

To generate diverse questions aligned with our taxonomy, we adopt a hybrid framework combining template, model, and human strategies:
(1) {Template-based (L1).} For atomic tasks, we design 45+ rule-based templates operating on MCGs to test grounding and attribute consistency.  
(2) {Model-assisted (L2).} For integration tasks, GPT-4o is prompted with structured instructions to generate questions involving cross-modal or temporal reasoning, refined by human reviewers.  
(3) {Human-authored (L3).} For reasoning tasks requiring multi-step inference, experts manually craft questions targeting compositional challenges like intention deception or rule violations. 
All questions undergo expert validation to ensure semantic rigor. The final benchmark contains 14.69k QA pairs balanced across three levels. A detailed composition breakdown is shown in Figure~\ref{fig:data_stat}.

\subsection{Quality Verification}
\label{ssec:quality_verfication}
In order to ensure the reliability of CrossCheck-Bench, we implement verification pipeline for addressing difficulty labeling, logic consistency, and robustness. Tasks are labeled as L1 (Perception), L2 (Integration), or L3 (Reasoning) via model consensus (76\% agreement across GPT-4o, GPT-4.1, Gemini 2.5 Pro) and finalized by expert annotators under a clear rubric. Experts override model votes in 18\% of cases.

We perform adversarial validation on 40\% of samples: three reviewers assess each QA pair for ambiguity or shortcuts, with 12\% flagged for revision. Inter-annotator agreement (IAA) on a separate 10\% confirms annotation stability. These steps ensure that CrossCheck-Bench is both challenging and reliable for multimodal reasoning evaluation.

\section{Experiments and Analysis}

\subsection{Setup}

We evaluate 13 vision-language models (VLMs), including both proprietary and open-source systems capable of processing interleaved image-text inputs. The proprietary models comprise Gemini 2.5 Pro\cite{comanici2025gemini}, GPT-4.1~\cite{achiam2023gpt}, GPT-4.1mini~\cite{achiam2023gpt}, and ByteDance-seed-1.6~\cite{seed2025seed1_5vl}. The open-source group spans multiple families and scales: Qwen2.5-VL(7B, 32B, 72B)\cite{bai2025qwen2}, InternVL3(38B, 78B)\cite{zhu2025internvl3} , Ovis2(8B, 16B, 34B)\cite{lu2024ovis}, MiMo-VL-7B\cite{xiaomi2025mimo}. All models are evaluated on CrossCheck-Bench using a unified zero-shot QA protocol with standardized prompts. Evaluation employs hybrid scoring: deterministic single-choice questions are assessed via exact match, while open-ended responses are semantically judged by GPT-4o. Open-source models are executed using official implementations on NVIDIA H100 GPUs. Human baseline performance is collected from seven expert annotators following the same QA protocols.

\begin{table*}[h]
\renewcommand{\arraystretch}{0.9}
\centering
\begin{tabular}{llccccc|c}
\toprule
\textbf{Task} & \textbf{Capacity} & GPT-4.1 & InternVL3-78B & Qwen2.5-VL-72B & Ovis2-34B & MiMo-VL-7B & \textbf{Overall-A(\%)} \\
\midrule
BR & A2 & 89.9 & 73.9 & 69.5 & 71.6 & 93.0 & 75.9 \\
BR* & \textbf{+ A4} & 80.1 ($\downarrow$9.8) & 74.0 ($\uparrow$0.1) & 69.1 ($\downarrow$0.4) & 69.0 ($\downarrow$2.6) & 66.1 ($\downarrow$26.9) & 71.5 ($\downarrow$4.4) \\
\midrule
SPU & A1,3 & 90.4 & 81.2 & 86.7 & 85.3 & 76.2 & 81.2 \\
SKU & \textbf{+ A5} & 72.1 ($\downarrow$18.3) & 71.3 ($\downarrow$9.9) & 68.3 ($\downarrow$18.4) & 66.2 ($\downarrow$19.1) & 56.6 ($\downarrow$19.6) & 64.4 ($\downarrow$16.8)\\
\midrule
BDC & A1,2,6 & 98.1 & 91.9 & 90.8 & 89.8 & 91.5 & 89.6 \\
PI & \textbf{+ A7} & 75.7 ($\downarrow$22.4) & 81.5 ($\downarrow$10.4) & 72.6 ($\downarrow$18.2) & 82.7 ($\downarrow$7.1) & 61.4 ($\downarrow$30.1) & 73.8 ($\downarrow$15.8) \\
BI & \textbf{+ A3,7} & 74.7 ($\downarrow$23.4) & 78.8 ($\downarrow$13.1) & 63.6 ($\downarrow$27.2) & 63.0 ($\downarrow$26.8) & 59.3 ($\downarrow$32.2) & 66.0 ($\downarrow$23.6) \\
BI* & \textbf{+ A3,4,7} & 65.4 ($\downarrow$32.7) & 64.0 ($\downarrow$27.9) & 69.2 ($\downarrow$21.6) & 55.9 ($\downarrow$33.9) & 46.7 ($\downarrow$44.8) & 59.0 ($\downarrow$30.6) \\
\bottomrule
\end{tabular}

\caption {Accuracy comparison on atomic vs. compositional tasks.Atomic variants and corresponding compositional variants are paired per task. Values in parentheses denote accuracy drops due to capability addition, measured relative to the baseline.}

\label{tab:atomic-vs-compositional}
\end{table*}

\subsection{Overall Results}
Table~\ref{tab:overall} summarizes the performance of 13 VLMs across 15 benchmark tasks, categorized into three task levels: perception, integration, and reasoning.

\noindent
\textbf{Human vs. MLLMs}  
Human accuracy provides an upper bound across all task levels. The average score reaches 95.2\%, surpassing the best proprietary model by over 18 points. Even on reasoning tasks, human performance remains above 88\%, while most models fall short of 76\%. This gap persists across categories, confirming that current models struggle with consistent multimodal alignment, particularly under conflicting or compositional input.

\noindent
\textbf{Open vs. Closed Models}  
Proprietary models consistently outperform open baselines. GPT-4.1 and Gemini 2.5 Pro achieve average scores above 76\%, while the strongest open-source model peaks at 71.5\%. Although open models perform competitively on localized tasks, their advantage vanishes as task complexity increases. The gap widens on composition-intensive queries, where proprietary systems maintain a consistent lead.

\noindent
\textbf{Performance Declines from Anchoring to Reasoning}  
Across all models, accuracy consistently declines as tasks progress from perceptual anchoring (L1) to knowledge integration (L2) and conflict reasoning (L3). Closed-source models show a clear downward trend, with GPT-4.1 dropping from 85.3\% on L1 tasks to 75.7\% on L3. Open-source models exhibit even sharper declines. InternVL3-78B falls from 71.5\% at L1 to 64.0\% at L3, and Qwen2.5-VL-72B drops from 69.9\% to 63.9\%. This pattern reveals that while most models can extract atomic entities and perform attribute-level comparisons, they struggle with reasoning over conflicting cues and enforcing logical consistency. The most severe degradation occurs in tasks that require multi-attribute fusion and rule-based contradiction detection, confirming that compositional reasoning remains the most fragile capability across architectures.

\noindent
\textbf{Model Scaling Yields Uneven Benefits Across Levels}  
Scaling improves performance on low-level tasks but brings inconsistent or diminishing returns at higher levels. For L1 tasks, larger models show notable gains. InternVL improves by 3.5 points from 38B to 78B, and Qwen2.5-VL improves by nearly 9 points from 7B to 72B. However, gains on L2 tasks are unstable. Qwen2.5-VL-72B improves marginally over 32B on some tasks, while degrading on others. At L3, reasoning accuracy stagnates or declines, even in large-scale models. These results indicate that increased capacity enhances surface perception but does not resolve the bottlenecks in cross-modal composition. The inability to align symbolic and visual information under constraint suggests that reasoning limitations persist independently of scale.

\subsection{Capability Analysis}

\paragraph{Atomic vs. Compositional Performance.}

Table~\ref{tab:atomic-vs-compositional} presents a controlled comparison between atomic and compositional tasks, revealing how capability integration impacts model reliability. While models handle isolated tasks (e.g., A1–A3) reasonably well, accuracy drops by 12\%–35\% when tasks involve numerical plausibility (A5), cross-frame reasoning (A4), or rule compliance (A7).

This compositional collapse is most evident when all capabilities (A1–A7) are required: top models often fail to exceed 50\% accuracy, with smaller ones below 40\%. Failures correlate strongly with tasks involving temporal (A4), numerical (A5), or rule-based (A7) reasoning, underscoring persistent architectural blind spots. CrossCheck-Bench disentangles these challenges, enabling precise attribution of integration failures.

\begin{figure}[t]
  \includegraphics[width=1.02\linewidth]{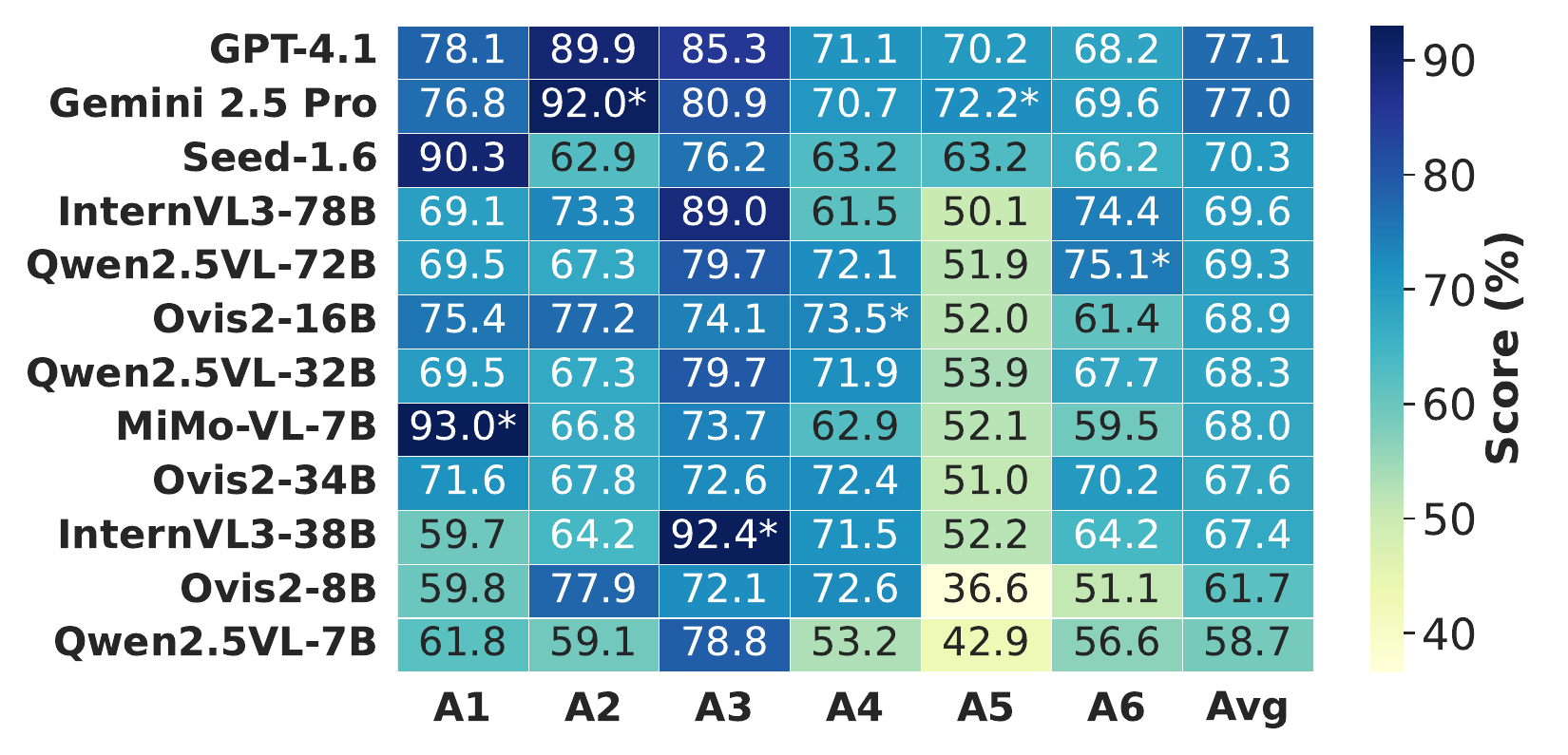}
  \caption{Model-wise Performance on Atomic Capabilities. Each cell indicates accuracy on one capability, with * denoting the best-performing model per capability.}
  \label{fig:atomic-skill-heatmap}
\end{figure}

\paragraph{Capability-Specific Trends and Scaling Effects.}

Figure~\ref{fig:atomic-skill-heatmap} summarizes model accuracy on atomic capabilities A1 through A6. These span from perceptual anchoring to symbolic reasoning. A clear separation emerges between perceptual skills (A1–A3) and reasoning-oriented capabilities (A4–A6). Models improve steadily on A1–A3 as scale increases. GPT-4.1 exceeds 85\% on A2 and A3. These skills rely on spatial anchoring and shallow visual-textual mapping, which align well with current pretraining objectives.
In contrast, A4 through A6 remain challenging across model families. Even top-tier models score below 75\% on average across these three capabilities. Smaller models collapse entirely. Ovis2-8B drops to 36.6\% on A5, and most open models perform below 55\% on A6. These failures suggest persistent fragility in symbolic inference when it depends on temporal alignment, numerical estimation, or rule-based logic.

Model scaling does not resolve these weaknesses uniformly. InternVL3-38B outperforms GPT-4.1 on A3, and Qwen2.5-VL-72B achieves the best A6 result overall. This suggests that architectural tuning or supervision strategies may contribute more than parameter count for certain atomic skills. While perceptual capabilities scale smoothly, symbolic coordination remains an unsolved frontier in vision-language alignment.

\subsection{Prompting-Based Diagnostic Interventions}
\label{ssec:prompting_interventions}

We test whether prompting or light tuning can recover capability-specific weaknesses revealed by CrossCheck-Bench. We focus on three difficult skills: numeric reasoning (A5), region-based OCR (A6), and logic inference (A7), where most models show low accuracy.

\begin{table}[t!]
\centering
\setlength{\tabcolsep}{4pt}
\begin{tabular}{lccc}
\toprule
\textbf{Intervention} & \textbf{A5: Numeric} & \textbf{A6: OCR} & \textbf{A7: Logic} \\
\midrule
Base (Vanilla) & 61.2 & 58.7 & 49.1 \\
CoT Prompting & 62.0 & 56.3 & \underline{50.8 $\uparrow$} \\
SoM Prompting & 62.4 & \underline{60.9 $\uparrow$} & 48.6 \\
CoT + SoM & 61.8 & 59.3 & 50.1 \\
CSFT & \underline{63.5 $\uparrow$} & 60.2 & 49.5 \\
MM-CoT & \textbf{65.3 $\uparrow$} & \textbf{61.7 $\uparrow$} & \textbf{53.5 $\uparrow$} \\
\bottomrule
\end{tabular}
\caption{Accuracy (\%) under prompting interventions on three capabilities. Gray cells mark best results. Bold and $\uparrow$ indicate top-2 gains over base.}
\label{tab:prompting_results}
\end{table}

\begin{figure}[t]
  \centering
  \includegraphics[width=\linewidth]{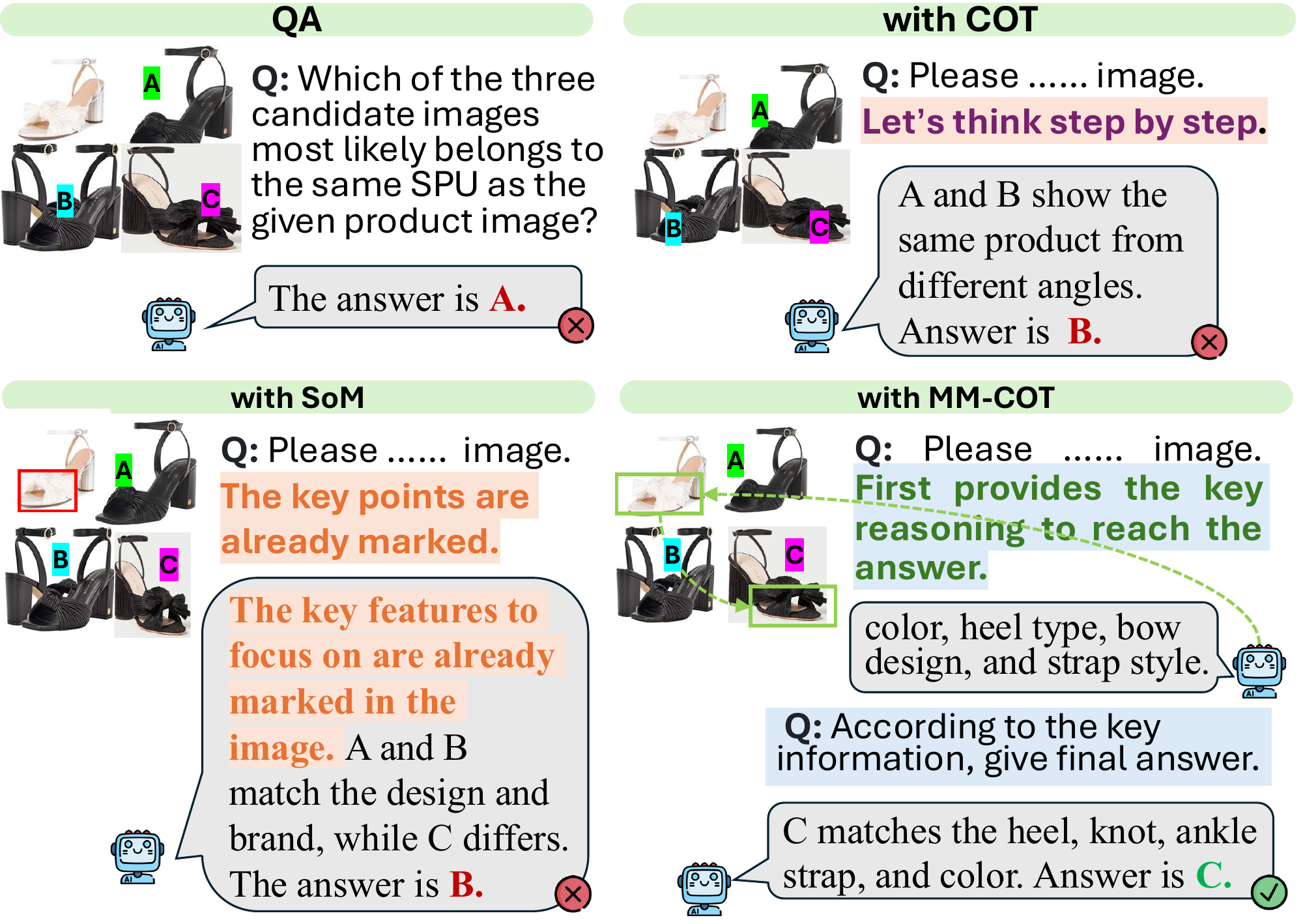}
  \caption{Prompting Strategies. We visualize answer rationales and visual focus under three prompting configurations. MM-CoT coordinates reasoning and grounding, leading to correct identification through structured inference.}
  \label{fig:prompting_case}
\end{figure}

\paragraph{Prompting and Fine-tuning Configurations.} We evaluate vision-language models under four strategies: Chain-of-Thought (CoT) prompting\cite{wei2022chain}, Set-of-Mark (SoM) visual guidance \cite{yang2023set}, a combined CoT+SoM setting, and supervised fine-tuning (CSFT) with 500 curated QA pairs. CoT adds symbolic reasoning scaffolds. SoM uses bounding boxes to guide attention. The combined setup merges both. CSFT tests model adaptability with lightweight supervision.

As shown in Table~\ref{tab:prompting_results}, CoT improves A7 performance (+1.7\% on average) but often harms perception-centric tasks due to hallucinated logic paths. SoM is particularly effective on A6 tasks for proprietary models with stronger visual pretraining, though results vary in open models. Combining CoT and SoM yields no consistent improvement and may introduce modality interference. CSFT helps moderately on A5 and A6 but fails to rectify A7 reasoning failures, suggesting tuning alone cannot repair logic abstraction gaps.

\paragraph{Multimodal Interleaved CoT (MM-CoT)}
To further address the failures, we propose \textit{Multimodal Interleaved CoT (MM-CoT)}, a two-stage prompting protocol designed to weave together grounding and reasoning. In Stage 1, models generate candidate answers with free-form rationales, which are parsed to extract relevant visual elements and highlight them using bounding-box overlays. In Stage 2, the model is re-invoked with both the SoM-augmented input and its own previous reasoning trace. This procedure encourages iterative inference, linking visual localization and symbolic logic through an explicit feedback loop. 
Figure~\ref{fig:prompting_case} shows a representative case under four prompting variants. 

\begin{figure}[t]
  \centering
  \includegraphics[width=\linewidth]{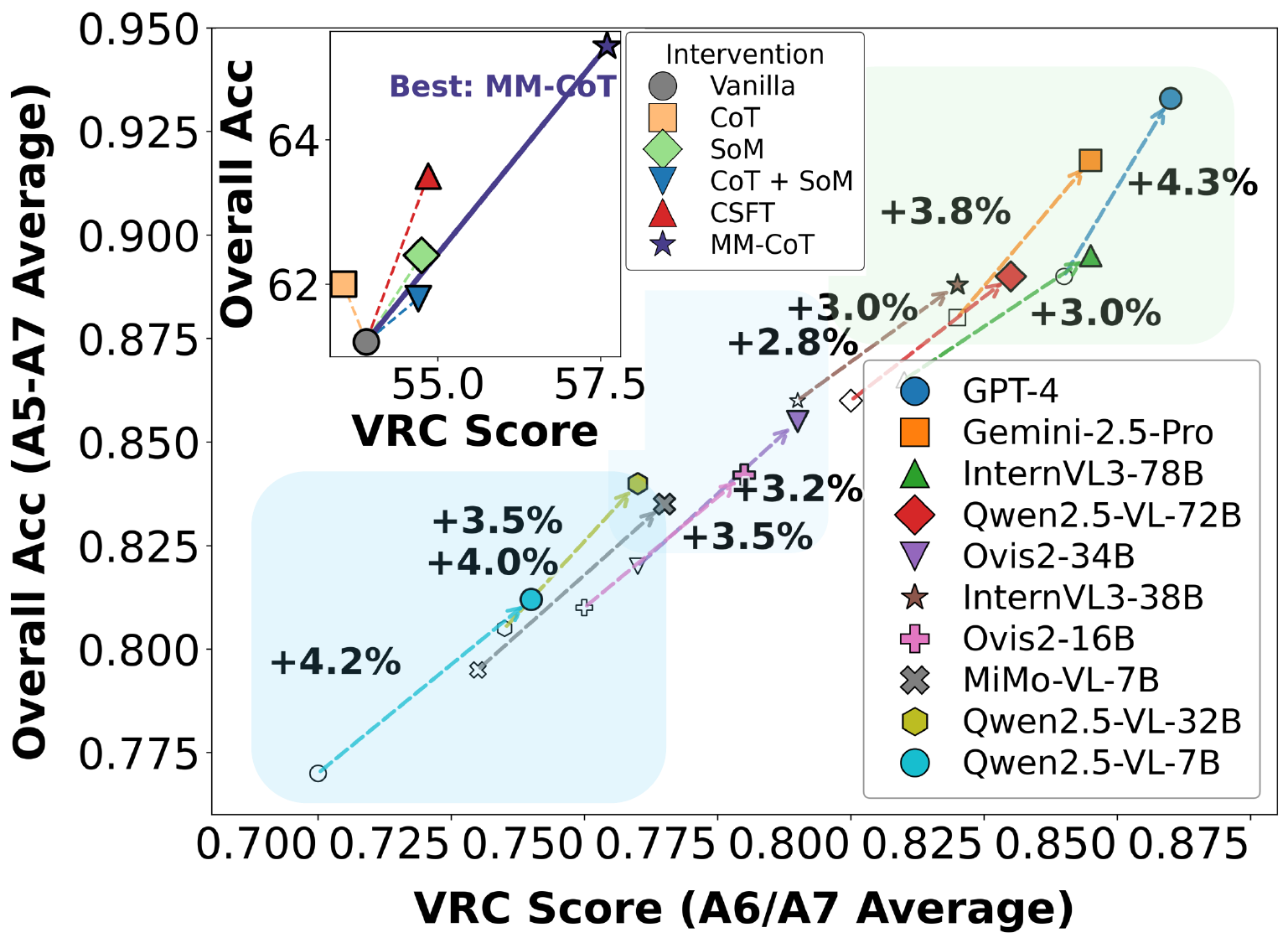}
  \caption{Effect of MM-CoT. The figure illustrates the performance shift across models with different prompting interventions. MM-CoT consistently enhances reasoning alignment and overall accuracy. Inset: comparison of intervention strategies on reasoning-intensive tasks (A5–A7).}
  \label{fig:mmcot_effect}
\end{figure}

As illustrated in Figure~\ref{fig:mmcot_effect}, MM-CoT consistently shifts models toward higher reasoning accuracy and VRC scores, especially for tasks involving compositional numerical and rule-based inference.
MM-CoT outperforms all prior strategies: GPT-4o sees a +4.4\% gain over vanilla prompting, and open models average +2.1\% improvement. Benefits are most significant in tasks demanding chained reasoning over values and constraints, such as those involving both A5 and A7. These findings demonstrate that tightly interleaved cross-modal prompting offers a viable path for addressing complex multimodal failures.

Overall, these results show that prompting strategies can selectively recover model deficits but struggle under compositional complexity. MM-CoT, by encouraging reasoning-grounding feedback loops, represents a promising direction for enhancing cross-modal inference---especially for tasks where visual and symbolic information must be reconciled through structured reasoning.

\section{Conclusion}
In this paper, we introduce CrossCheck-Bench, a diagnostic benchmark for evaluating vision-language models (VLMs) under multimodal inconsistency. The benchmark targets models' ability to detect conflicts and reason compositionally across visual, textual, and symbolic inputs.
CrossCheck-Bench defines a three-tier capability structure and spans 15 tasks covering atomic and compositional reasoning. Through systematic evaluation, we uncover three key trends: (1) performance degrades consistently from perception to reasoning; (2) models prioritize internal priors over conflicting external signals; (3) tasks requiring cross-frame alignment and rule-grounded inference remain major bottlenecks.
These failures are not random but traceable to structural capability gaps. By isolating where compositional failures emerge, CrossCheck-Bench offers a roadmap for improving alignment between symbolic inference and grounded perception. 
Additional probing shows that conventional prompting strategies such as Chain-of-Thought and Set-of-Mark yield only marginal gains. By contrast, methods that interleave symbolic reasoning with grounded visual processing achieve more stable improvements. 
We hope this work enables more robust, reliable, and cognitively grounded VLMs for real-world applications, and provides a foundation for future research on multimodal consistency.

\bibliography{aaai2026} 

\clearpage

\section{Appendix}
\setcounter{figure}{7}
\setcounter{table}{3} 
\section*{Appendix A \quad Atomic Capability Definitions}
\addcontentsline{toc}{section}{Appendix A \quad Atomic Capability Definitions}

This appendix provides detailed definitions of the atomic capabilities that underpin \textsc{CrossCheck-Bench}. Each capability corresponds to a fundamental multimodal skill required to resolve inconsistencies across visual and textual modalities. The definitions below outline the cognitive operation involved, followed by a representative task example. Table~\ref{tab:atomic-capabilities} summarizes these definitions in a compact reference format.

\paragraph*{A1 \, Spatial Anchoring}
Identifies and localises the most relevant object or region within an image. This capability forms the foundation for grounding subsequent reasoning in the correct visual context. While seemingly simple, challenges arise when salient entities are visually ambiguous, occluded, or embedded in clutter.
\noindent\textit{Example task:} detecting the core product in a retail image.

\paragraph*{A2 \, Character Recognition}
Extracts alphanumeric text from images, including brand names, product tags, or serial numbers. Unlike general OCR, this capability prioritises accurate text understanding under varying font styles, layouts, or partial occlusions.
\noindent\textit{Example task:} extracting a brand name from stylised product packaging.

\paragraph*{A3 \, Attribute Comparison}
Compares visual attributes—such as logo shape, color, or design—across multiple entities. This includes both intra-frame and inter-frame comparisons, and is critical for assessing consistency among visually similar instances.
\noindent\textit{Example task:} determining whether two logos belong to the same brand.

\paragraph*{A4 \, Cross-modal Alignment}
Determines whether visual and textual modalities refer to the same real-world entity. This requires mapping semantics across modalities—e.g., linking a brand name in text to a logo in an image—and detecting mismatches.
\noindent\textit{Example task:} verifying that a product image matches its title description.

\paragraph*{A5 \, Value Plausibility Estimation}
Assesses whether numerical or categorical values inferred from the image are contextually reasonable. This capability involves applying implicit world knowledge or domain-specific heuristics to judge values for plausibility.
\noindent\textit{Example task:} flagging a luxury bag priced at \$1 as suspicious.

\paragraph*{A6 \, Region-Constrained OCR}
Performs OCR within a tightly defined image region, typically specified by a prior visual anchor (e.g., bounding box). Unlike general text recognition, this task requires spatial precision and the ability to disambiguate target zones.
\noindent\textit{Example task:} reading embedded brand text from the main part of a product.

\paragraph*{A7 \, Rule Compliance Reasoning}
Executes rule-based inference tasks grounded in predefined policies or logical constraints. This includes identifying violations of branding guidelines, marketplace rules, or platform-specific listing policies.
\noindent\textit{Example task:} validating that a product listing does not violate intellectual property rules based on visual and textual evidence.

\begin{table}[h!]
  \centering
  \small
  \setlength{\tabcolsep}{4pt}
  \captionsetup{position=bottom}

  \begin{tabular}{
    c
    >{\RaggedRight}m{2.9cm}
    >{\RaggedRight}m{2.3cm}
    l
  }
    \toprule
    \textbf{ID} & \textbf{Dataset (Abbr.)} & \textbf{Task} & \textbf{Capabilities} \\
    \midrule
    1 & LogoDiff (LD)          & Logo detail comparison   & A3 \\
    2 & BrandRecognition (BR)  & Product brand recognition & A2 \\
    3 & ProductMainPart (PMP)  & Product main-part localisation & A1 \\
    4 & ProductMainPartOCR (PMO) & Main-part OCR & A6 \\
    5 & ProductDiff\textsuperscript{*} (PD*) & Cross-frame product match & A4 \\
    6 & StylePrice (SP)        & Style price recognition   & A5 \\
    \bottomrule
  \end{tabular}

  \caption{L1 datasets and assessed capabilities.}
  \label{tab:l1-datasets}
\end{table}

\begin{table}[h!]
  \centering
  \small
  \setlength{\tabcolsep}{4pt}
  \captionsetup{position=bottom}

  \begin{tabular}{
    c
    >{\RaggedRight}m{3.0cm}
    >{\RaggedRight}m{2.3cm}
    l
  }
    \toprule
    \textbf{ID} & \textbf{Dataset (Abbr.)} & \textbf{Task} & \textbf{Capabilities} \\
    \midrule
    7  & BrandRecognition\textsuperscript{*} (BR*) & Content brand recognition & A2 + A4 \\
    8  & SPU  & Same-style SPU match & A1 + A3 \\
    9  & SKU  & Same-style SKU match & A1 + A3 + A5 \\
    10 & BrandCategory (BDC) & Brand operating category & A1 + A2 + A6 \\
    \bottomrule
  \end{tabular}

  \caption{L2 datasets and assessed capabilities.}
  \label{tab:l2-datasets}
\end{table}

\begin{table}[h!]
  \centering
  \small
  \setlength{\tabcolsep}{4pt}
  \captionsetup{position=bottom}

  \begin{tabular}{
    c
    >{\RaggedRight}m{2.1cm}
    >{\RaggedRight}m{2.1cm}
    >{\RaggedRight}m{2.5cm}
  }
    \toprule
    \textbf{ID} & \textbf{Dataset (Abbr.)} & \textbf{Task} & \textbf{Capabilities} \\
    \midrule

    11 & Brand-Circumvention (BC)
       & Product logo occlusion
       & A1 + A6 + A7 \\

    12 & Brand-Circumvention\textsuperscript{*} (BC*)
       & Content logo occlusion
       & A1 + A4 + A6 + A7 \\

    13 & BrandIntent (BI)
       & Brand intent (4-class)
       & A1 + A2 + A3 + A6 + A7 \\

    14 & ProductIntent (PI)
       & Product intent recognition
       & A1 + A2 + A6 + A7 \\

    15 & BrandIntent\textsuperscript{*} (BI*)
       & Content brand intent (4-class)
       & A1 + A2 + A3 + A4 + A6 + A7 \\

    \bottomrule
  \end{tabular}

  \caption{L3 datasets and their required atomic capabilities.}
  \label{tab:l3-datasets}
\end{table}

\vspace{2mm}
\begin{table*}[t]
  \centering
  \small
  \setlength{\tabcolsep}{5pt}
  \renewcommand{\arraystretch}{1.07}

  \begin{tabular}{
    p{0.55cm}      
    p{3.2cm}       
    p{7.2cm}       
    p{4.4cm}       
  }
    \toprule
    \textbf{ID} & \textbf{Capability} & \textbf{Definition} & \textbf{Representative Benchmark Task} \\
    \midrule

    A1 & Spatial Anchoring &
        Identify and localise the primary object or region of interest in an image, grounding visual attention for downstream reasoning. &
        Detecting the core product in a cluttered retail display. \\

    A2 & Character Recognition &
        Extract and interpret alphanumeric text from visual input, including stylised or partially occluded text. &
        Extracting brand names from product packaging. \\

    A3 & Attribute Comparison &
        Compare fine-grained visual attributes across entities—such as logo shape, texture, or colour—to assess visual consistency. &
        Determining whether two logos refer to the same brand. \\

    A4 & Cross-modal Alignment &
        Judge whether visual and textual inputs semantically refer to the same object or concept. &
        Verifying image–text consistency in a product listing. \\

    A5 & Value Plausibility Estimation &
        Evaluate whether numerical or categorical values inferred from content are contextually plausible. &
        Identifying suspicious prices that indicate possible fraud. \\

    A6 & Region-Constrained OCR &
        Extract text from a predefined spatial region within the image, requiring spatial accuracy and focus. &
        Reading brand names from the designated core zone of a product. \\

    A7 & Rule Compliance Reasoning &
        Apply explicit logical or policy-based rules to detect violations or inconsistencies. &
        Flagging listings that violate intellectual property policies. \\

    \bottomrule
  \end{tabular}

  \caption{Atomic capability definitions with representative benchmark tasks.}
  \label{tab:atomic-capabilities}
\end{table*}

\section*{Appendix B\quad Dataset--Capability Mapping}
\addcontentsline{toc}{section}{Appendix B \quad Dataset--Capability Mapping}

This appendix details how atomic capabilities are instantiated across the 15 datasets in \textsc{CrossCheck-Bench}. To enable fine-grained analysis, we group tasks into three levels of cognitive complexity: \textbf{Perception (L1)}, \textbf{Integration (L2)}, and \textbf{Reasoning (L3)}, each emphasizing progressively richer combinations of multimodal skills. For each level, we summarize the corresponding datasets and the subset of atomic capabilities they are designed to evaluate.

\paragraph*{Level 1 – Perception}

This level focuses on basic perceptual skills that form the foundation of multimodal understanding. Tasks require models to localize objects, recognize text, compare low-level visual attributes, or extract structured information. All tasks are grounded in visually salient cues and avoid intermodal composition.

\paragraph*{Level 2 – Integration}
Integration tasks combine two or more atomic skills to test cross-modal or cross-entity alignment. These tasks move beyond perception by requiring models to reconcile information across textual and visual modalities (e.g., matching product names with visual attributes), or across similar entities (e.g., comparing SKUs/SPUs). Although still grounded in surface-level inputs, successful completion now demands consistent attribute mapping and multimodal referent resolution.

\paragraph*{Level 3 – Reasoning}
Reasoning-level tasks require holistic understanding, often involving partial evidence, occluded clues, or symbolic rules. These tasks incorporate higher-order capabilities such as intent classification, contextual text parsing, and rule compliance checks. Unlike L1–L2 tasks, success in L3 hinges on the ability to integrate multiple clues, resolve contradictions, and simulate inference patterns akin to human judgment.

\section*{Appendix C \quad Prompt Templates}
\addcontentsline{toc}{section}{Appendix D \quad Prompt Templates and Input Formats}

All evaluations in \textsc{CrossCheck-Bench} follow a unified vision-language QA format that interleaves image(s) and textual input. Each prompt is designed to test one or more atomic capabilities, with inputs presented either as single-image QA, multi-view QA, or image-text contradiction detection. This section outlines the generic prompting format and provides representative templates.

\paragraph{Unified QA Format.}
Each QA instance is structured as:
\begin{quote}
\texttt{<User>:} Given the following input, answer the question. \\
\texttt{<Image(s)>} + \texttt{<Text>} + \texttt{<Question>} \\
\texttt{<Assistant>:} [free-form answer or option choice]
\end{quote}

For classification-style tasks (e.g., intent recognition), the answer space is predefined. For open-ended responses, we rely on semantic scoring.

\paragraph{Example 1: Product–Title Consistency (A4).}
\texttt{<Image>:} [image of a red running shoe] \\
\texttt{<Text>:} Title: "Men's Leather Oxford Dress Shoes" \\
\texttt{<Question>:} Does the image match the product title?

\paragraph{Example 2: Cross-frame Product Match (A1 + A3).}
\texttt{<Image 1>:} [side-view of a bag] \\
\texttt{<Image 2>:} [front-view of a similar-looking bag] \\
\texttt{<Question>:} Do both images depict the same product?

\paragraph{Example 3: Brand Text Extraction (A2 + A6).}
\texttt{<Image>:} [cropped product logo region] \\
\texttt{<Question>:} What brand name shown in the region?

\paragraph{Prompting Strategies.}
\begin{itemize}
    \item {Chain-of-Thought (CoT):} appended instruction “Explain your reasoning step by step.”
    \item {Set-of-Mark (SoM):} visual regions highlighted with bounding boxes in image input.
\end{itemize}

\section*{Appendix D \quad Model Inference Protocol}
\addcontentsline{toc}{section}{Appendix E \quad Model Execution and Inference Protocol}

To ensure fairness and reproducibility, we standardize inference protocols across all evaluated models. This section details the execution strategy, input modality handling, and decoding parameters used.
\paragraph{Decoding Parameters.}
All models use the following settings unless otherwise specified:
\begin{itemize}
    \item \texttt{temperature = 0.7}, \texttt{top\_p = 0.9}, \texttt{max tokens = 512}
    \item {Image input:} resized and normalized per model specs (e.g., 224px or 448px short edge)
    \item {Image format:} JPEG or PNG with bounding box markup when needed (e.g., for A6 tasks)
\end{itemize}

\paragraph{Output Evaluation.}
\begin{itemize}
    \item {Closed QA tasks} are scored via exact match.
    \item {Open-ended tasks} are semantically scored by GPT-4o with rubric guidance.
\end{itemize}
To reduce variance, all outputs are cached and postprocessed uniformly (lowercased, stripped, unicode normalized).

\section*{Appendix E\quad Human Annotation Protocol and Quality Assurance}
\addcontentsline{toc}{section}{Appendix F \quad Human Annotation Protocol and Quality Assurance}

The benchmark includes 15k carefully curated question-answer pairs derived from real-world artifacts. This appendix outlines the human annotation workflow, annotation toolchain, and quality control process.

\paragraph{Annotation Pipeline.}
Annotation proceeds in three sequential stages:
\begin{enumerate}
    \item \textbf{Ground Truth Collection:} Source product pages, metadata, and OCR outputs are parsed to construct initial candidate QA entries.
    \item \textbf{Synthetic Contradiction Injection:} Visual or textual inconsistencies (e.g., swapped prices, mismatched brands) are inserted to simulate real-world errors.
    \item \textbf{Expert Review:} Annotators verify QA integrity, confirm capability coverage, and ensure scenario realism.
\end{enumerate}

\paragraph{Annotator Pool.}
All annotators were trained graduate-level researchers or domain experts with prior experience in V+L tasks or content moderation. Each task required 2 annotators and 1 verifier.

\paragraph{Annotation Interface.}
We developed a custom interface allowing side-by-side image comparison, bounding box drawing, and free-form QA creation. Task-specific instructions were embedded to reduce ambiguity.

\paragraph{Quality Assurance.}
\begin{itemize}
    \item \textbf{Double annotation:} Each sample reviewed by two annotators independently.
    \item \textbf{Inter-annotator agreement:} On a stratified 10\% subset, we achieved 92.6\% agreement with Cohen’s $\kappa = 0.88$.
    \item \textbf{Difficulty Labeling:} Each task was assigned an L1/L2/L3 difficulty label during annotation, followed by multi-round cross-batch consistency checks to maintain labeling reliability and reduce annotation drift.
    \item \textbf{Spot auditing:} Weekly sampling and manual spot-checking by senior reviewers ensured consistency.
\end{itemize}

In total, over \textbf{450 human-hours} were devoted to benchmark construction and validation.

\section*{Appendix F\quad Human Baseline Evaluation}

Seven expert annotators with prior experience in visual content moderation, product attribution, or e-commerce content analysis were tasked with the same instructions, QA protocols, and annotation interface as used in dataset construction, ensuring their judgments are directly comparable to model outputs.

Each annotator independently completed the benchmark, covering all capability categories (A1–A7). For tasks involving open-ended rationales or multi-step reasoning (e.g., cross-modal alignment or rule-compliance assessment), annotators were instructed to provide short textual justifications to confirm that their decisions aligned with the task definitions.

To measure the consistency and reliability of human judgments, we computed inter-annotator agreement (IAA) using task-appropriate metrics. Human baseline performance is then reported as the mean score across annotators, with variance reflecting annotation difficulty and inherent task ambiguity. This human benchmark provides an empirical upper bound for model comparison and a reference for evaluating task complexity.

\section*{Appendix G\quad Task Taxonomy}
\begin{figure*}[h]
    \centering 
    \includegraphics[width=\textwidth]{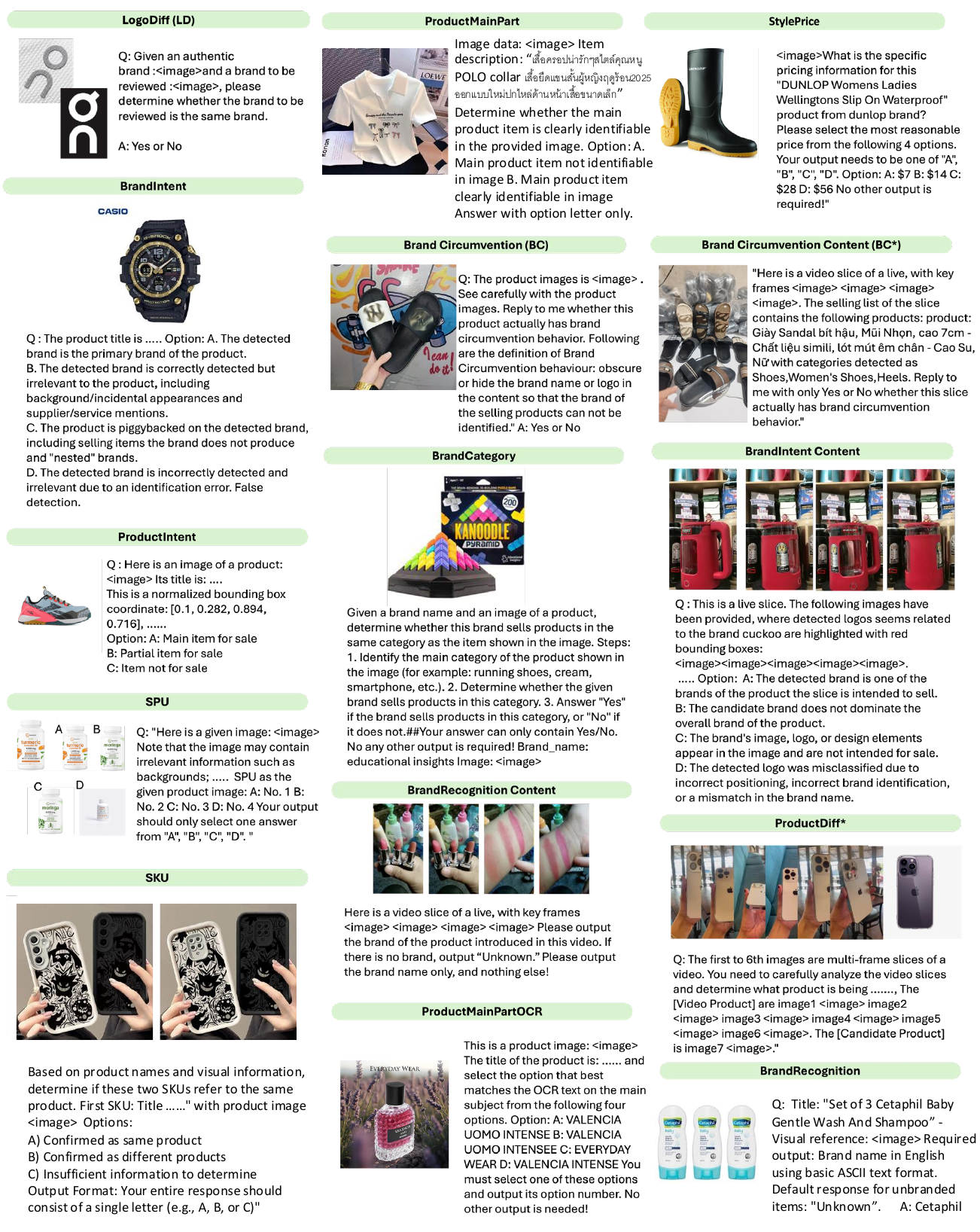}
    \caption{Overview of all 15 benchmark tasks in CrossCheck-Bench, represented as image-text combinations. Each task is categorized based on its cognitive level and capability composition, with examples illustrating the variety of input modalities (e.g., single image, image pair, or image-text combination) and the corresponding output format. }
    \label{fig:Taxonomy}
\end{figure*}

To illustrate the scope and compositional diversity of \textsc{CrossCheck-Bench}, we provide a visual taxonomy of all 15 benchmark tasks, grouped by cognitive level and capability composition. Each task is represented with a real or representative example, highlighting its input modality (e.g., single image, image pair, or image-text combination) and output format.
Figure~\ref{fig:Taxonomy} summarizes these examples.

\end{document}